\documentclass[a4paper,french]{rnti} 

\usepackage[T1]{fontenc}
\usepackage[utf8]{inputenc}

\usepackage{url}
\usepackage{graphicx}
\usepackage{color}

\usepackage{placeins}
\usepackage{multirow}
\usepackage{listings}
\usepackage{caption}
\usepackage{subcaption}
\usepackage{hhline}

\titrecourt{Amélioration de la qualité d'images}

\titre{Amélioration de la qualité d'images avec un algorithme d'optimisation inspirée par la nature}

\nomcourt{O: Parisot et al.}
\auteur{Olivier Parisot, Thomas Tamisier}
\affiliation{Luxembourg Institute of Science and Technology (LIST) \\ 5 Avenue des Hauts-Fourneaux L-4422 Luxembourg \\ olivier.parisot@list.lu}

\resume{
L'application de prétraitements explicites et reproductibles est fondamentale dans le domaine de la vision par ordinateur, pour pouvoir comparer efficacement des algorithmes ou pour préparer un nouveau corpus d'images.
Dans cet article, nous proposons une méthode pour obtenir une séquence reproductible de transformations qui améliore une image donnée: le calcul est réalisé via un algorithme d'optimisation inspirée par la nature et basé sur des techniques d'évaluation de la qualité.
Des tests montrent l'impact de l'approche sur différents ensembles d'images de l'état de l'art. 
}

\summary{
Reproducible images preprocessing is important in the field of computer vision, for efficient algorithms comparison or for new images corpus preparation.
In this paper, we propose a method to obtain an explicit and ordered sequence of transformations that improves a given image: the computation is performed via a nature-inspired optimization algorithm based on quality assessment techniques.
Preliminary tests show the impact of the approach on different state-of-the-art data sets. 
}

\begin{document}

L'amélioration de la qualité des images est un sujet d'intérêt dans le domaine de la vision par ordinateur.
De nouvelles techniques sont régulièrement proposées afin de traiter efficacement le flou, le bruit, les problèmes de contraste et même la faible résolution (\cite{parekh2021survey}).
Ainsi, la sélection et l'application de transformations appropriées est une tâche complexe qui nécessite un savoir-faire et des outils spécifiques pour le traitement d'images.
Dans un premier temps, il est essentiel d'utiliser des métriques pour guider le processus: à cet égard, l'évaluation de la qualité des images (\emph{IQA -- Image Quality Assesment}) vise à estimer automatiquement la qualité d'une image de manière à ce qu'elle corresponde à une évaluation humaine subjective (\cite{zhai2020perceptual}).

Par ailleurs, en recherche scientifique et en ingénierie, il est plus que jamais utile voire nécessaire de garantir la reproductibilité des expériences sur les jeux de données à base d'images en disposant de la trace des transformations effectuées (\cite{berg2018progress}).
Un travail récent a montré qu'une information complète et rigoureuse au sujet des traitements effectués sur les images fait défaut dans une grande proportion d'articles scientifiques, ce qui peut compromettre l'interprétation des résultats obtenus (\cite{miura2021reproducible}).
Nous pouvons faire une analogie avec l'apprentissage automatique lorsqu'il est appliqué sur des données numériques ou catégorielles: le prétraitement des données doit être documenté de manière précise pour aboutir à des modèles prédictifs significatifs et de confiance (\cite{zelaya2019towards}).

L'approche que nous présentons procède par optimisation inspirée de la nature pour améliorer une image: elle s'appuie sur des méthodes standard d'évaluation de la qualité et elle construit la trace explicite de la séquence de transformations successivement opérées.
Le reste de cet article est organisé comme suit. 
Tout d'abord, les travaux connexes sur l'évaluation de la qualité et l'optimisation inspirée par la nature pour les images sont brièvement présentés (Section \ref{sec:related}). 
Ensuite, notre approche pour améliorer la qualité des images est introduit (Section \ref{sec:approach}). 
Un prototype est décrit (Section \ref{sec:implementation}) et les résultats des expériences réalisées sont discutés (Section \ref{sec:experiences}).
Pour finir, nous concluons en ouvrant quelques perspectives (Section \ref{sec:conclusion}).

\section{Etat de l'art}
\label{sec:related}

\subsection{Evaluation de la qualité d'image}

De nombreuses approches d'évaluation de la qualité d'image ont été développées ces dernières années et une liste exhaustive a déjà été compilée par \cite{zhai2020perceptual}. 
On peut distinguer deux grandes familles de techniques: 
\begin{itemize}
\item Les méthodes à référence complète (Full Reference IQA -- FR-IQA) et à référence réduite (Reduced Reference IQA -- RR-IQA) sont basées sur un référentiel d'images (brutes/déformées). 
\item Les techniques sans référence (No Reference IQA -- NR-IQA) et à l'aveugle (Blind-IQA) visent à estimer la qualité de manière non supervisée (\cite{liu2019no}).
\end{itemize} 

Dans cet article, nous nous concentrons sur l'utilisation d'approches NR-IQA et Blind-IQA et parmi lesquelles nous pouvons citer:
\begin{itemize}
	\item BRISQUE (Blind/Referenceless Image Spatial Quality Evaluator) proposé par \cite{mittal2012no}: un score entre 0 et 100 est produit (0 pour une image de bonne qualité, 100 pour une qualité médiocre) .
	\item NIMA (Neural Image Assessment) introduit par \cite{talebi2018nima}: un ensemble de modèles de Deep Learning pour estimer la qualité esthétique et technique des images: un score entre 0 et 10 est produit (0 pour une image de mauvaise qualité, 10 pour une image de bonne qualité).
\end{itemize}

En pratique, ces méthodes d'évaluation sont largement utilisées dans les benchmarks pour comparer l'efficacité des algorithmes de traitement d'images, comme décrit par \cite{li2018benchmarking}.
Elles permettent de détecter avec une efficacité certaine les images de mauvaise qualité.

\subsection{Optimisation inspirée par la nature pour le traitement d'images}

A l'instar des réseaux neuromimétiques, de nombreuses approches de résolution de problèmes ou d'optimisation sont inspirées par des processus naturels, notamment les algorithmes évolutionnaires et l'optimisation par essaims de particules (\cite{li2020newly}).
Ces techniques sont fréquemment appliquées en vision par ordinateur pour diverses tâches telles que la réduction du flou et du bruit, la restauration et la segmentation (\cite{dhal2019survey,ramson2019nature}).
En particulier, \cite{pal1994genetic} propose un algorithme génétique pour mettre en valeur automatiquement une image en la modifiant de manière progressive.

Dans la suite de cet article, nous proposons une solution pour la génération de séquences ordonnées de transformations d'images en appliquant avec un algorithme guidé par les techniques d'évaluation de la qualité des images précédement citées.

\section{Approche}
\label{sec:approach}

L'élément central de l'approche est défini de la manière suivante:
\begin{itemize}
	\item Une image initiale.
	\item Une séquence ordonnée de transformations appliquée sur l'image initiale (exemples: ajustement du contraste puis suppression du brouillard, suivi d'un ajustement de l'histogramme et d'un débruitage par variation totale, etc.). Au départ, cette séquence est vide.
	\item Un score de qualité est évalué avec les méthodes d'évaluation de la qualité de l'image BRISQUE ou NIMA (esthétique ou technique). Cette étape est critique et pilote l'algorithme (la qualité sert ici de \emph{fitness} de la solution, dans la terminologie utilisée pour les algorithmes évolutionnaires).
\end{itemize}

Pour une image d'entrée donnée (I), en considérant une méthode d'évaluation de la qualité de l'image (M) et un nombre maximum d'époques (E), l'algorithme suivant calcule un ensemble de séquences ordonnées de transformations à appliquer sur l'image: 
\begin{itemize}
	\item Une population est générée avec P images: chaque image est un clone de l'image initiale I sur laquelle une transformation aléatoire a été appliquée ou non. En effet, pour s'assurer que l'algorithme ne conduit pas à une image de moindre qualité, il est important de conserver au moins un clone non modifié de l'image initiale dans la population dans le cas où l'algorithme ne produit pas de meilleure solution.
	\item Pendant E itérations :
	\begin{itemize}
		\item La meilleure image présente dans la population ainsi qu'une autre image choisie au hasard sont clonées, puis une transformation aléatoire est appliquée pour chacune: les images nouvellement créées sont évaluées avec M et ajoutées à la population.
		\item Une autre image est sélectionnée aléatoirement dans la population et est empilée avec l'image initiale (avec un poids aléatoire): l'élément nouvellement créé est évalué avec M et ajouté à la population.
		\item Selon l'évaluation avec M des images présentes dans la population, les plus mauvaises images sont sélectionnées puis retirées de la population (pour toujours garder P images dans la population).
	\end{itemize}
	\item Le résultat final est l'image de la population consolidée ayant la meilleure estimation de qualité. La sortie de l'algorithme est alors une séquence ordonnée de transformations qui conduit à une amélioration de l'évaluation de la qualité de l'image.
\end{itemize}

Pour éviter que l'image ne diffère trop de l'image initiale, nous avons ajouté un test comparant la similarité entre l'image produite et l'image initiale: si la similarité est trop faible (c'est-à-dire inférieure à un seuil prédéfini T), alors le score de l'image est fortement pénalisé et la dernière transformation appliquée n'est donc pas considérée comme acceptable.
Le test est ici basé sur l'indice de similarité structurelle (Structural Similarity Index): en pratique, la valeur est proche de 1 lorsque les deux images sont similaires alors que la valeur est proche de 0 lorsque les images sont vraiment différentes.

\section{Implementation et expériences}
\label{sec:implementation}

L'algorithme a été implémenté dans un prototype Python.
Divers composants open-source standards ont été intégrés.
Le chargement et la transformation des images sont réalisés à l'aide de divers packages Python dédiés tels que \emph{openCV} \footnote{\url{https://pypi.org/project/opencv-python/}} et \emph{scikit-images}. \footnote{\url{https://pypi.org/project/scikit-image/}}.
Le score BRISQUE est calculé grâce au package \emph{image-quality} \footnote{\url{https://pypi.org/project/image-quality/}} et les scores NIMA sont fournis par une implémentation de Tensorflow \footnote{\url{https://github.com/idealo/image-quality-assessment}}.

En utilisant ces composants, les transformations d'images suivantes peuvent être appliquées et combinées à la volée via notre prototype:
\begin{itemize}
	\item Débruitage via différentes méthodes: variation totale, moyennes non locales, ondelettes, bilatéral.
	\item Ajustement du contraste et de la luminosité notamment via des algorithmes comme CLAHE (Contrast Limited Adaptive histogram equalization, \cite{clahe1994}). 
	\item Traitement du \emph{background}, notamment pour améliorer les images à faible luminosité (\cite{8769288}).
	\item Atténuation du brouillard via des modèles de Deep Learning comme Cycle-Dehaze (\cite{engin2018cycle}).
	\item Transformations morphologiques comme erosion et dilatation (\cite{sreedhar2012enhancement}).
	\item Reconstruction des détails avec des modèles comme Noise2Noise (\cite{lehtinen2018noise2noise}).
	\item Empilage pondéré et/ou soustraction d'images.
\end{itemize}

Ce code Python a été développé de manière à pouvoir être exécuté sur une infrastructure haute performance ayant la configuration matérielle suivante: 40 cœurs et 128 Go de RAM (CPU Intel(R) Xeon(R) Silver 4210 @ 2,20 GHz) et NVIDIA Tesla V100-PCIE-32 Go.
Plus précisémment, les librairies CUDA \footnote{\url{https://developer.nvidia.com/cuda-zone}} et NUMBA \footnote{\url{http://numba.pydata.org/}} ont été utilisées autant que possible pour accélérer l'exécution de code sur cette infrastructure.

\section{Expériences}
\label{sec:experiences}

\begin{figure}
	\centering
	\begin{subfigure}{.329\textwidth}
		\centering
		\includegraphics[width=.99\linewidth]{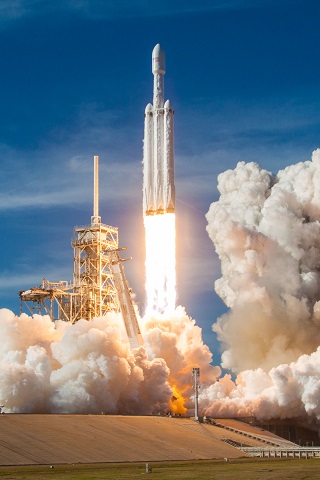}
		\caption{Image originale, aucune modification apportée: BRISQUE=9.637 et NIMA(esthét.)=5.166.} 
		\label{fig:res10}
	\end{subfigure}
	\begin{subfigure}{.329\textwidth}
		\centering
		\includegraphics[width=.99\linewidth]{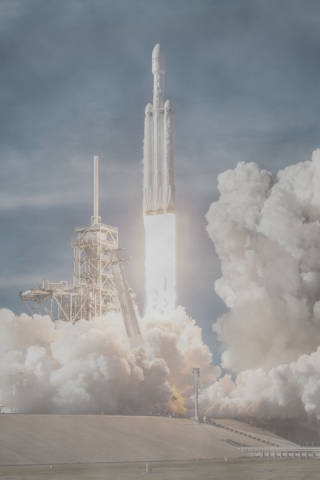}
		\caption{Image artificiellement dégradée avec l'ajout de brouillard: BRISQUE=14.771 et NIMA(esthét.)=5.071.}
		\label{fig:res11}
	\end{subfigure}
	\begin{subfigure}{.329\textwidth}
		\centering
		\includegraphics[width=.99\linewidth]{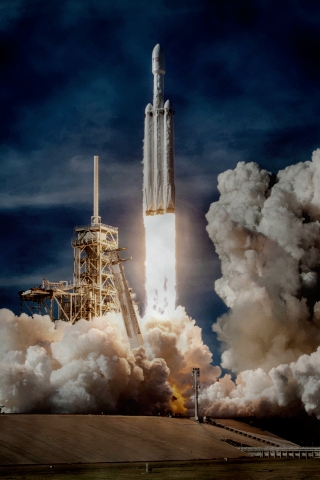}
		\caption{Résultat après application de notre algorithme sur l'image dégradée: BRISQUE=3.134 et NIMA(esthét.)=5.160. }
		\label{fig:res12}
	\end{subfigure}
	\begin{subfigure}{.329\textwidth}
		\centering
		\includegraphics[width=.99\linewidth]{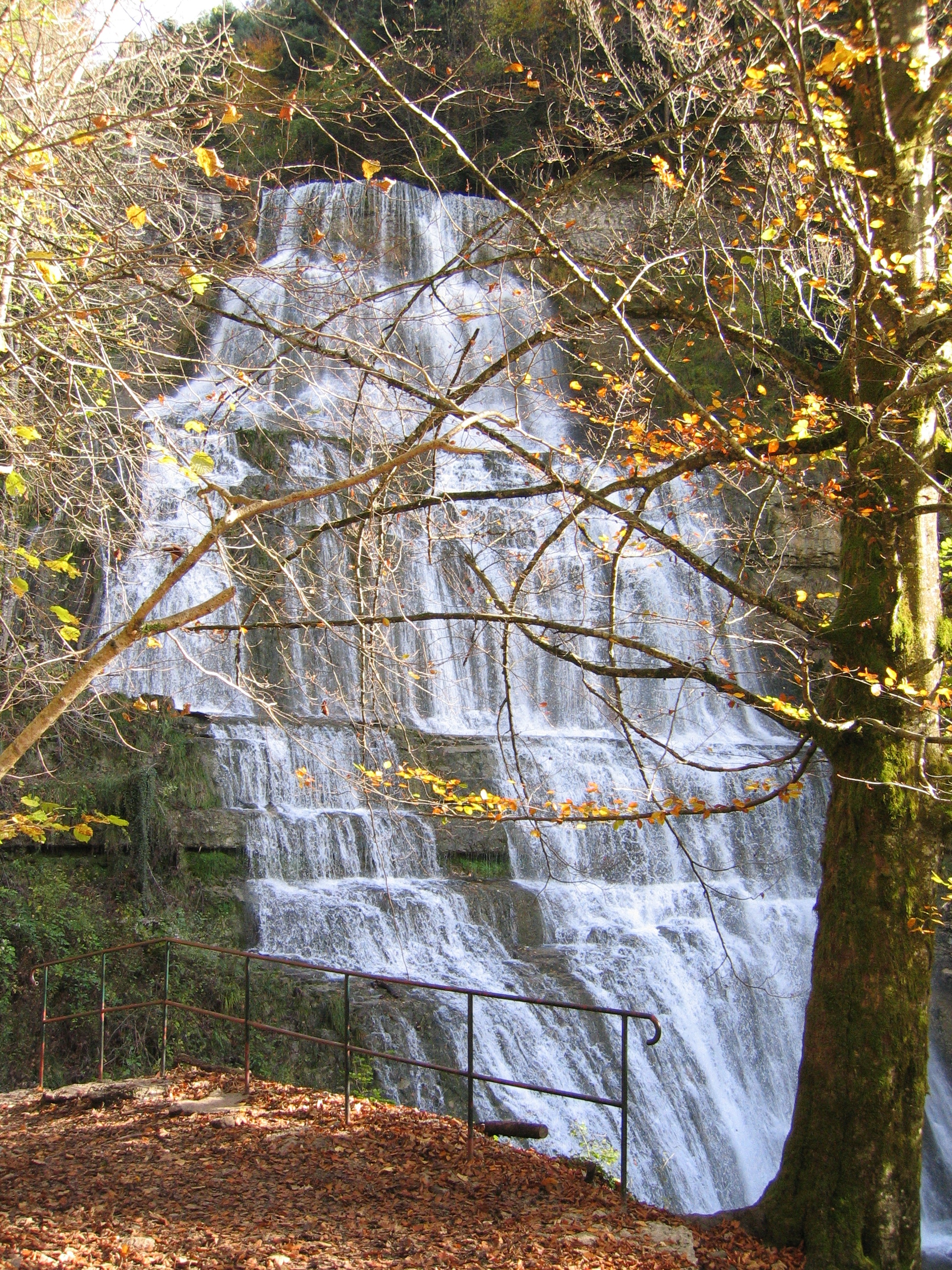}
		\caption{Image originale, aucune modification apportée: BRISQUE=21.999 et NIMA(esthét.)=5.105.} 
		\label{fig:res20}
	\end{subfigure}
	\begin{subfigure}{.329\textwidth}
		\centering
		\includegraphics[width=.99\linewidth]{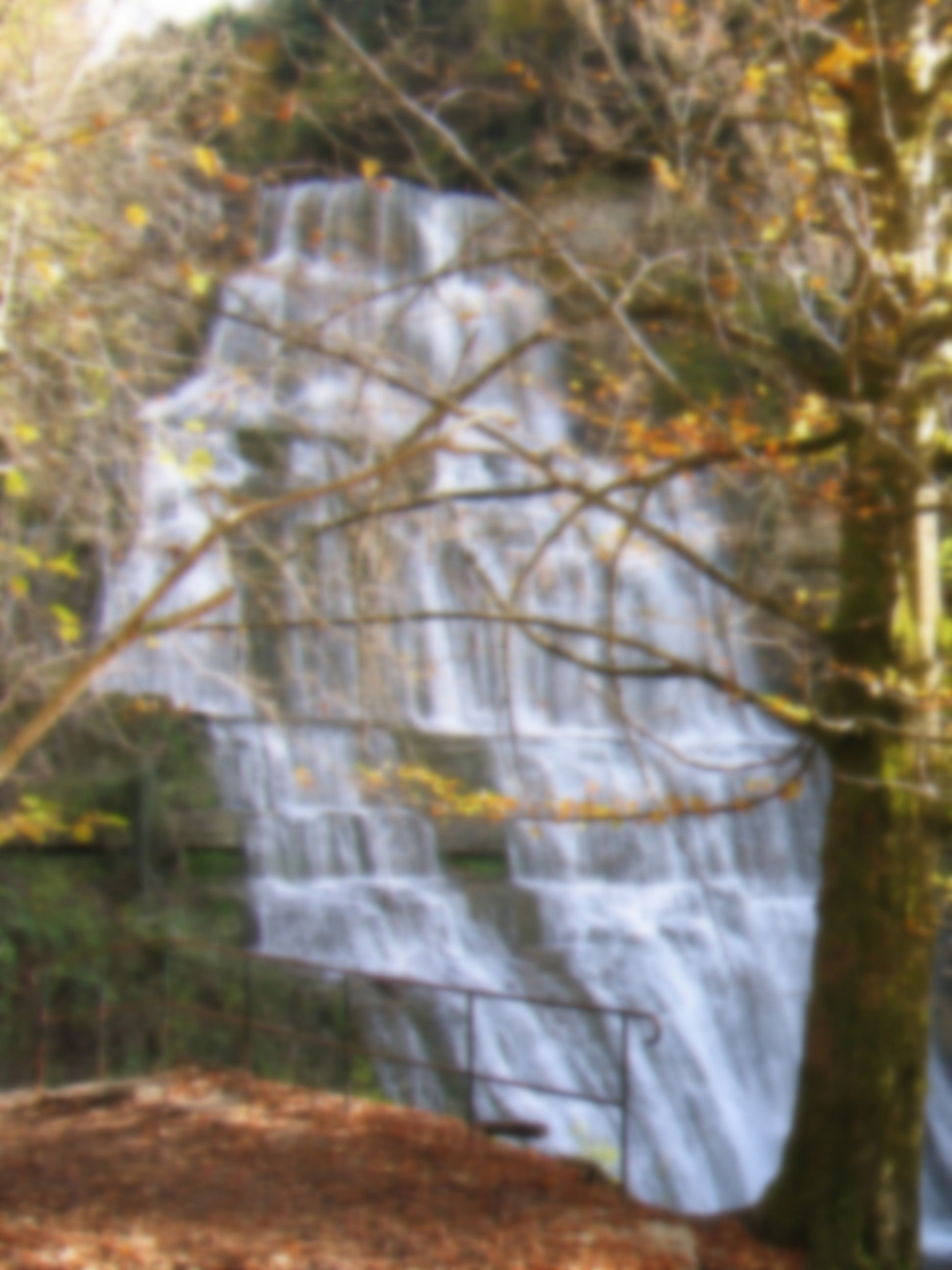}
		\caption{Image artificiellement dégradée avec l'ajout de flou: BRISQUE=69.136 et NIMA(esthét.)=3.749.}
		\label{fig:res21}
	\end{subfigure}
	\begin{subfigure}{.329\textwidth}
		\centering
		\includegraphics[width=.99\linewidth]{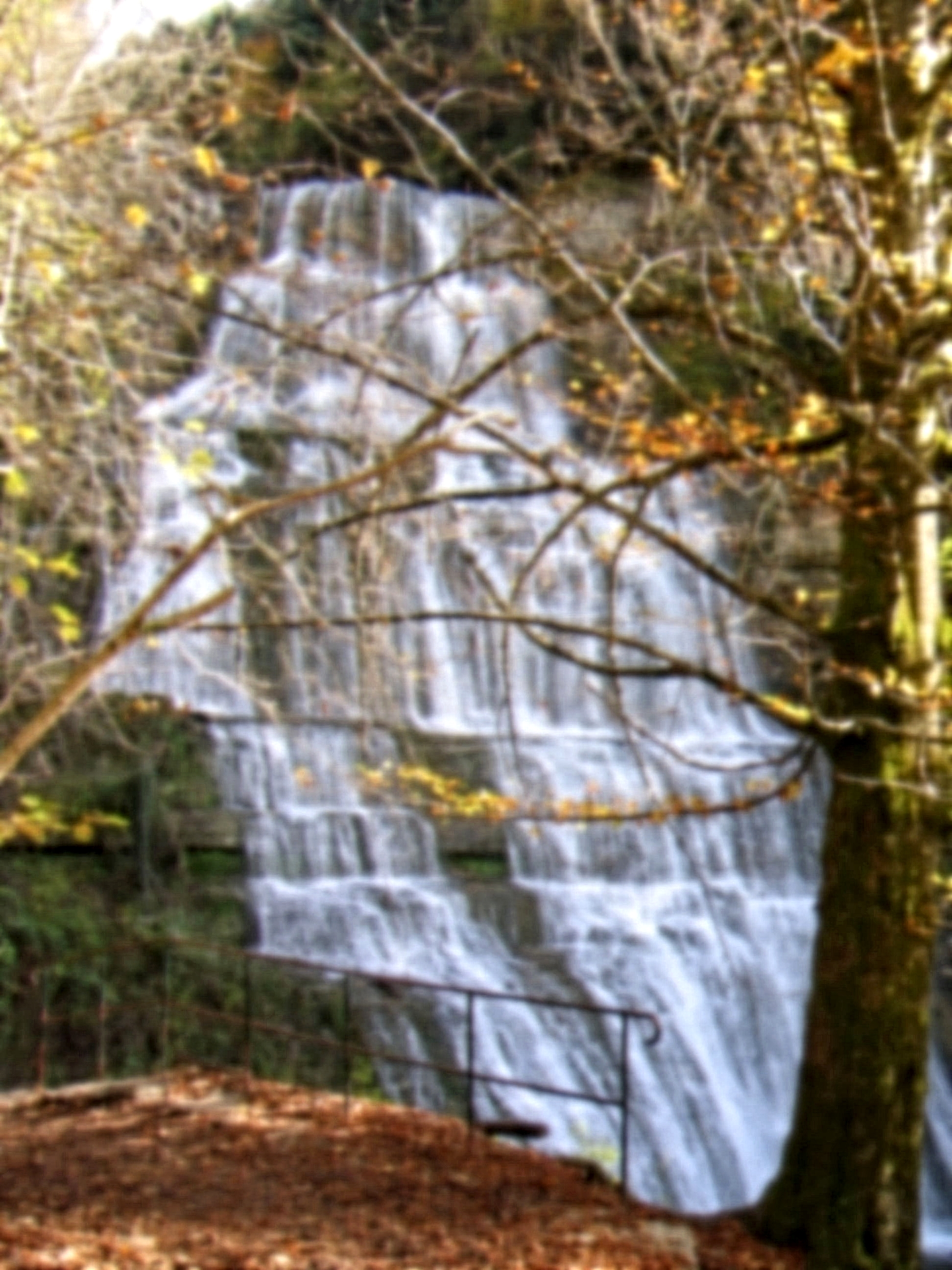}
		\caption{Résultat après application de notre algorithme sur l'image dégradée: BRISQUE=33.544 et NIMA(esthét.)=3.896. }
		\label{fig:res22}
	\end{subfigure}
	\caption{En haut: une image de la fusée Falcon Heavy de l'entreprise SpaceX (\ref{fig:res10}), une version artificiellement dégradée de l'image (\ref{fig:res11}) et cette dernière traitée par notre algorithme (\ref{fig:res12}). En bas: une photo de la Cascade du Hérisson (France, Jura) (\ref{fig:res20}), une version artificiellement dégradée de la photo (\ref{fig:res21}) et cette dernière traitée par notre algorithme (\ref{fig:res22}).}
	\label{fig:results2}
\end{figure}

L'algorithme a d'abord été testé sur des photographies telles que Figure \ref{fig:results2} \footnote{\url{https://commons.wikimedia.org/wiki/Main_Page}}.
Dans l'exemple de l'image de la fusée, les transformations suivantes ont été appliquées sur une version volontairement altérée de la photographie initiale: \emph{désembrumage, empilement avec image initiale (avec poids à 0.8), débruitage par filtrage bilateral, augmentation du contraste, traitement de la netteté (avec beta à 0.95), augmentation du contraste, débruitage par filtrage bilateral, suppression du fond, augmentation du contraste, traitement de la netteté (beta: 1.05)}.
Cette séquence ordonnée d'opérations permet d'améliorer la qualité mesurée de l'image dégradée (en se référant aux scores BRISQUE et NIMA).

Dans un second temps, plusieurs jeux de données de référence en traitement d'images ont été considérées durant nos expérimentations: TID2013 avec 500 images déformées choisies au hasard (\cite{ponomarenko2015image}) et LIVE avec 175 images bruitées (\cite{sheikh2006statistical}).
De plus, nous avons testé l'approche sur les benchmarks CID2013 et LOL: le premier contient 475 images capturées avec des appareils grand public comme les smartphones (\cite{cid2013}) et le second consiste en 485 images à faible luminosité (\cite{wei2018deep}).

Les résultats du Tableau \ref{table:results} ont été obtenus avec les hyperparamètres suivants: NIMA-esthétique comme score d'évaluation de la qualité d'image ciblée, une population initiale de 20 images, 50 époques maximum et 0,5 comme similarité minimale.
Selon nos différents essais, cette configuration de l'algorithme offre le meilleur compromis entre l'amélioration de la qualité et le temps d'exécution. 
Le score BRISQUE a été calculé par la suite pour estimer d'une autre manière et à posteriori la qualité des résultats de l'algorithme et la variance du bruit (Noise Variance) a été calculée pour mettre en évidence le niveau de bruit dans les benchmarks.

\begin{table}[]
	\centering
	\caption{Expériences sur plusieurs bases de données d'images de l'état de l'art: les métriques moyennes des évaluations de la qualité des images (avant et après l'exécution de l'algorithme) sont listées.}	
	\label{table:results}
	\begin{tabular}{|l|l|r|r|r|}
		\hline
		\multicolumn{2}{|l|}{Images benchmark}  		  & BRISQUE & NIMA (esthétique)  \\ \hline
		\multirow{2}{*}{TID2013}      			& avant & 45.271  & 4.952  		      \\  
		& après   & 34.168  & 5.792  		      \\ 
		\hline
		\multirow{2}{*}{LIVE} 		   			& avant & 59.309  & 5.005  		    \\ 
		& après   & 54.833  & 5.710  		     \\ 
		\hhline{|=|=|=|=|=|}
		\multirow{2}{*}{CID2013} 		   		& avant & 13.560  & 4.778  		    \\ 
		& après   & 28.990  & 5.277  		    \\ 
		\hline
		\multirow{2}{*}{LOL} 					& avant & 21.158  & 4.706  		       \\ 
		& après   & 51.241  & 5.464  		      \\ 
		\hline
	\end{tabular}
\end{table}

D'après ces premières expériences, les scores de qualité (NIMA-esthétique et BRISQUE) sont meilleurs pour les images provenant des benchmarks TID2013 et LIVE après l'exécution de notre algorithme.
Idem pour le bruit: notre algorithme tend à le réduire dans les images transformées pour ces benchmarks.

Nous notons cependant que les résultats pour les images à faible luminosité sont partiellement satisfaisants (benchmark LOL).
Même si les transformations calculées par notre algorithme augmentent le score NIMA-esthétique comme souhaité, elles ont tendance à dégrader le score BRISQUE et à augmenter le bruit.
Ce problème montre les limites de NIMA et BRISQUE pour l'évaluation de la qualité des images à faible luminosité: ces méthodes ne devraient pas être utilisées pour guider la transformation de telles images.

En ce qui concerne les performances de calcul, le temps nécessaire pour exécuter le prototype Python était raisonnable en utilisant l'infrastructure décrite ci-dessus (de quelques secondes à plusieurs dizaines de secondes par image, en fonction de leur résolution). 
Le temps d'exécution du prototype dépend pour l'essentiel de la méthode d'évaluation de la qualité appliquée: par exemple, l'évaluation NIMA-esthétique est réalisée en moyenne en moins de 100 millisecondes tandis que l'évaluation du score BRISQUE peut prendre plusieurs secondes.
En pratique, ce temps peut être réduit en diminuant la taille des images avant évaluation (mais cela affecte également la précision des évaluations, ce qui n'est pas forcément souhaitable).

\section{Conclusion and perspectives}
\label{sec:conclusion}

Cet article présente un algorithme d'optimisation inspirée de la nature améliorant la qualité mesurée d'une image avec le calcul d'une séquence ordonnée et reproductible de transformations à appliquer.
Un prototype Python basé sur les méthodes d'évaluation de la qualité des images a été mis en oeuvre et testé sur diverses bases de données d'images de l'état de l'art.

Divers partenaires académiques et opérationnels permettront de mettre en place des cas d'utilisation réels de l'approche (notamment en bio-informatique).
En parallèle, nous améliorons le prototype en générant automatiquement le code source Python permettant de transformer les images, en nous inspirant de ce qui est proposé dans les approches AutoML (\emph{Automated Machine Learning}).
De plus, les futurs cas d'utilisation seront également utilisés pour améliorer le prototype proposé sur les images à faible luminosité.
Enfin, les performances d'exécution seront renforcées en distribuant les calculs via des frameworks comme Spark.
\\

\section*{Remerciements}

Ce travail a été réalisé en utilisant la plateforme \textit{Artificial Intelligence, Data Analytics} opérée par le Luxembourg Institute of Science and Technology.
Dans ce cadre, nous remercions tout particulièrement Jean-François Merche et Raynald Jadoul pour leur support.

\bibliographystyle{rnti}
\bibliography{article}

\Fr

\end{document}